# Small Target Detection for Search and Rescue Operations using Distributed Deep Learning and Synthetic Data Generation

Kyongsik Yun*, Luan Nguyen, Tuan Nguyen, Doyoung Kim,
Sarah Eldin, Alexander Huyen, Thomas Lu, Edward Chow
Jet Propulsion Laboratory, California Institute of Technology
4800 Oak Grove Drive, Pasadena, CA 91109

## ABSTRACT

It is important to find the target as soon as possible for search and rescue operations. Surveillance camera systems and unmanned aerial vehicles (UAVs) are used to support search and rescue. Automatic object detection is important because a person cannot monitor multiple surveillance screens simultaneously for 24 hours. Also, the object is often too small to be recognized by the human eye on the surveillance screen. This study used UAVs around the Port of Houston and fixed surveillance cameras to build an automatic target detection system that supports the US Coast Guard (USCG) to help find targets (e.g., person overboard). We combined image segmentation, enhancement, and convolution neural networks to reduce detection time to detect small targets. We compared the performance between the auto-detection system and the human eye. Our system detected the target within 8 seconds, but the human eye detected the target within 25 seconds. Our systems also used synthetic data generation and data augmentation techniques to improve target detection accuracy. This solution may help the search and rescue operations of the first responders in a timely manner.

## 1. INTRODUCTION

Unmanned aerial vehicles (UAVs) have long been considered for search and rescue operations [1–3]. One of the important advantages of UAVs used in search and rescue operations is that it is possible to detect small objects in a wide range when the victim is far away and difficult to find. Currently UAV video cameras have very high resolution (4k). It's hard to find the injured distant person in the high-definition video with the human eye. The human eye has the advantage of knowing the context of the image and finding where it is likely to find victims based on prior experiences. But the human eye can focus on only a small part at a time. Also, the object we are looking for in search and rescue operations is very small and usually in a dark environment.

Reducing time for search and rescue is important. In many cases, the survivability of the victim decreases exponentially over time [2,4]. To save time, we applied a combination of image segmentation, image enhancement, and convolutional neural networks to detect small humans using video recorded from UAV. Image segmentation is a patch-based system that uses divide-and-conquer strategies to partition images into overlapping patches and extracts the features of each patch [5]. This algorithm has been used to demonstrate improved object detection performance [5,6].

Previous studies developed a search plan to improve accuracy and latency to locate victims [7]. SARPlan, developed by the Canadian Forces to optimize the search mission, begins by defining the potential area with a geographic decision support system. Then, based on the effort availability and victim probability in the map, the system optimally allocates possible effort to generate an optimal search plan. Another related example is the EU-ICARUS project, which develops assistive robot tools for search and rescue operations [8]. Robot tools for sensing and perception collect data from multiple sensors installed in UAVs or external data sources (e.g., CCTV) and combine them for target detection and tracking. This research focused specifically on robotic tools and lacked sophisticated target recognition algorithms to operate them.

With the recent advances in deep neural network architectures, we achieved unprecedented target recognition accuracy and latency [9–11]. Transfer learning, incremental retraining, and data augmentation techniques have proven effective for target recognition [12]. DARPA announced the subterranean challenge last year, and one of its required capabilities is to detect hazardous objects in a dark, unknown environment [13]. Onboard deep neural network processing is required because of limited communication connectivity when underground. Technology development to support search and rescue operations is becoming increasingly important [14].

*kyun@jpl.nasa.gov; phone 1 818 354-1468; fax 1 818 393-6752; jpl.nasa.gov

This study used UAVs and fixed surveillance cameras to increase accuracy and shorten the time to recognize small objects (e.g., a person in the water, a person injured in the mountain).

## 2. DISTRIBUTED DEEP LEARNING

Detecting small objects in large images is a difficult problem. The size of the network input is limited by the amount of memory available on the GPU. The performance of deep learning-based object recognition degrades when small objects are detected. The reason is that using transfer learning, the system learns about general features from a pre-trained network, because most objects are relatively large in the image of a common public training data set. In a previous study, patch-level object detection was performed to show that this method achieved much higher detection accuracy than fast RCNN for small objects [6]. Another previous study used wavelet transforms to improve object detection performance in search and rescue operations [15].

We used a distributed deep learning system. The first step was to compartmentalize each part of high-definition video into many smaller images. The image was then preprocessed to improve contrast. We used multiple single shot detector (SSD) modules to detect the target object in the small image and recombine the small images into the original high-resolution image to get the final result (Figure 1). Our system detected the victim within 8 seconds of UAV-based search and rescue operations, but the human video analyst was able to detect the victim within 25 seconds, thus greatly improving the operational efficiency.

In this study, scale invariance was not considered due to computational efficiency. We assume that the target size is relatively constant (very small) because of the distance between the unmanned aircraft and the ground, and the distance between the CCTV and the target on the ground in search and rescue operations. Scale invariance should be considered for future generalized systems.

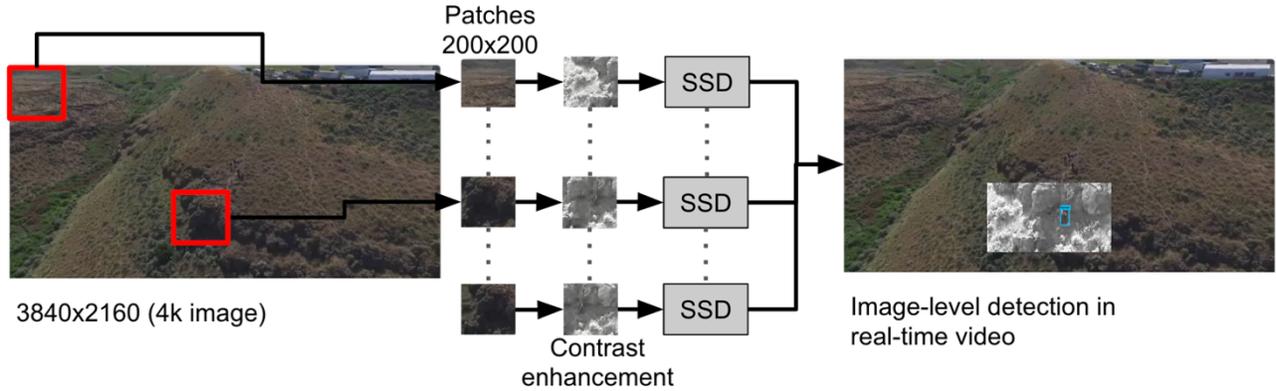

Figure 1. An illustration of the proposed framework. The original 4k video was divided into small patches (200x200) as inputs to single shot multibox detector (SSD) modules to produce patch-level object detection results. Image-level detection results were produced by projecting all patch-level results back to the original image.

## 3. SYNTHETIC DATA GENERATION

We created an image using the 3D game editor ARMA3 (Bohemia Interactive, Amsterdam, The Netherlands). ARMA3 is an open, realism-based, military tactical video game that delivers realistic synthetic image generation. ARMA3 is based on Real Virtuality 4 3D engine. ARMA3 uses NVIDIA PhysX, an open source real-time physics engine middleware SDK, to enable real-world simulations of vehicles, land, sea and air. We developed a script to work with ARMA3, which spawns random people and vehicles at random locations in a specified location and imports images from multiple angles in a given 3D environment. We can also specify the motion and behavior of the object. For example, a person can walk down a street or direct a truck to run on a specified route. Through this process, we created hundreds of high-quality images in a matter of hours. In reality, it can take days or weeks to take images in the real world. The relative cost of this synthetic data generation process is negligible compared to the potential cost of collecting similar data in real life.

We further processed the synthetic image using a selective Gaussian blur filter with a blur radius of 5 pixels and a maximum delta of 50. The maximum delta is the maximum difference between the pixel value and the surrounding pixel value (value between 0-255). Those pixels with values higher than this max delta value are not blurred (Figure 2). For the target detection test of search and rescue operations, we used a video database provided by Houston Port, Harris County, Texas and the US Coast Guard.

ARMA3 was used to generate a synthetic dataset of search and rescue operations. The synthetic dataset mimics the situation at various spatial perspectives, different times of the day, and background (Figure 2). In these various situations, creating an actual image can be costly and time-consuming. With a 3D game engine, however, it takes no more than 20 seconds to create a high-quality synthetic image.

With synthetic data, we can pinpoint the exact location of each object (such as a person, car, etc.) based on 3D model information, and place each object anywhere. We can also shuffle the background and the target to create an infinite combination of synthetic data. For target recognition and segmentation purposes, objects are already fully annotated in the 3D model and do not require manual labeling. In particular, manual labeling for segmentation requires a great deal of time and resources when using real images. Manual labeling by humans may interfere with effective training of neural networks by creating inconsistent ground truth data sets between the labels. The synthetic data solves all of these problems.

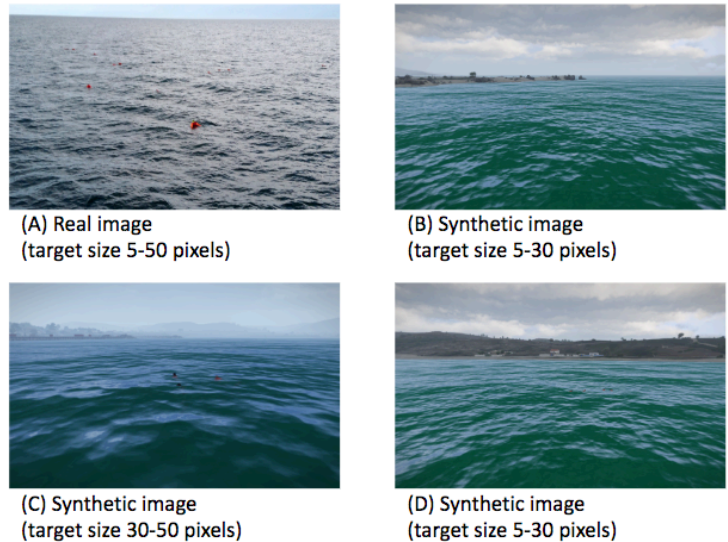

Figure 2. Representative man overboard images. (A) Real image, including near and far objects (target size 5-50 pixels), (B) synthetic image of far objects with partial ground view (target size 5-30 pixels), (C) synthetic image of near objects with foggy partial ground view (target size 30-50 pixels), (D) synthetic image of far objects with ground view (target size 5-30 pixels).

## 4. TRAINING AND TESTING

First, we used 97 actual images for training. The loss converged after 2000 iterations. We tested 109 real images for detection of small target victims in search and rescue operations and obtained the mean average precision (mAP) of 0.77 under intersection over union (IoU) of 0.5. We then trained the networks separately with 90 synthetic images and compared their performance against the same 109 real images for testing. We slightly improved the mAP from 0.77 to 0.79 in the same IoU. The result means that nearly identical object detection performance was achieved using real and synthetic data sets. This indicates that the synthetic dataset was as efficient as the real data for training.

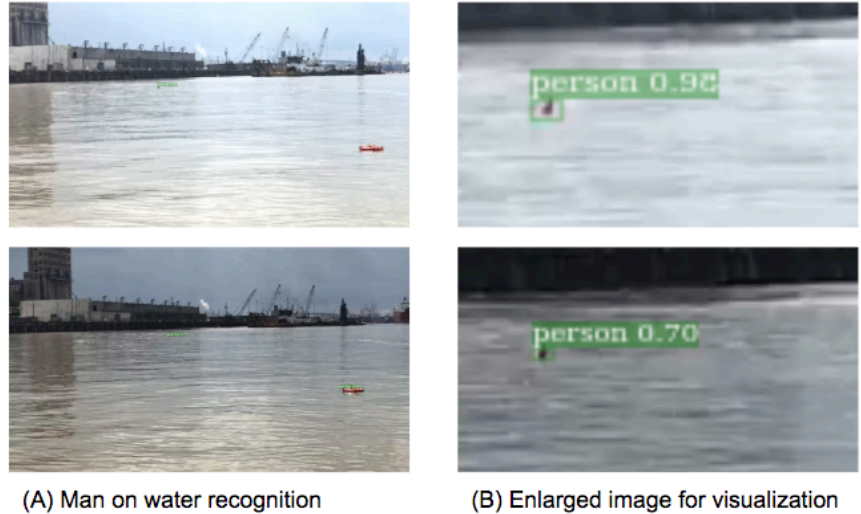

Figure 3. Representative man on water recognition testing results. (A) Man on water detection in Houston Port Operational Experimentation, (B) enlarged images of the target for visualization.

As we added more data to the system, the average precision and recall improved (Figure 4). In the training using 2029 real images, the average precision was 82.16% and the average recall was 76.83%. By adding 430 real images, 348 synthetic images generated from the 3D game engine, and 1529 traditionally augmented images by flipping, rotating, and zooming in/out, the results improved to the average precision of 84.06% and the average recall of 78.35%, respectively.

## 5. DISCUSSION

We built a distributed deep learning system to efficiently and accurately recognize small objects in long distances. We adopted synthetic data generation and data augmentation to train the system on a limited set of data. We found that the addition of synthetic and augmented data improved the accuracy of detection. While the system detected objects within 8 seconds of search and rescue operations, a human analyst detected objects within 25 seconds.

We used an SSD module for object recognition. Future studies can improve accuracy using a variation of SSD networks, including concatenation and element-sum of each layer. By fusing multi-level features, concatenation modules and element-sum modules can reduce interference caused by useless background noise and improve the importance of contextual information. Previous studies showed that this special module improved the object detection performance of the SSD[16].

We could further improve the search and rescue operations using multiple UAVs. Distributed deep learning can be implemented across multiple UAVs, considering video streams from multiple UAVs as one giant image. Combining UAV's sensor and image analysis processing can help optimize UAVs flight parameters, including position of the UAVs, energy constraints, environmental hazards, and data sharing constraints [2].

Our synthetic data generation method using ARMA3, a 3D game engine, can be effectively applied to automated image labeling to generate unlimited, high-quality training data that requires little human resources [17,18]. To improve accuracy, we should also consider generating procedural synthetic data [19]. If we look at only one frame of a given video sequence, we will not be able to detect small objects in many cases because of noise, such as lights, obstructions, and waves in search and rescue operations. However, looking at the relationship between frames makes target object detection more accurate.

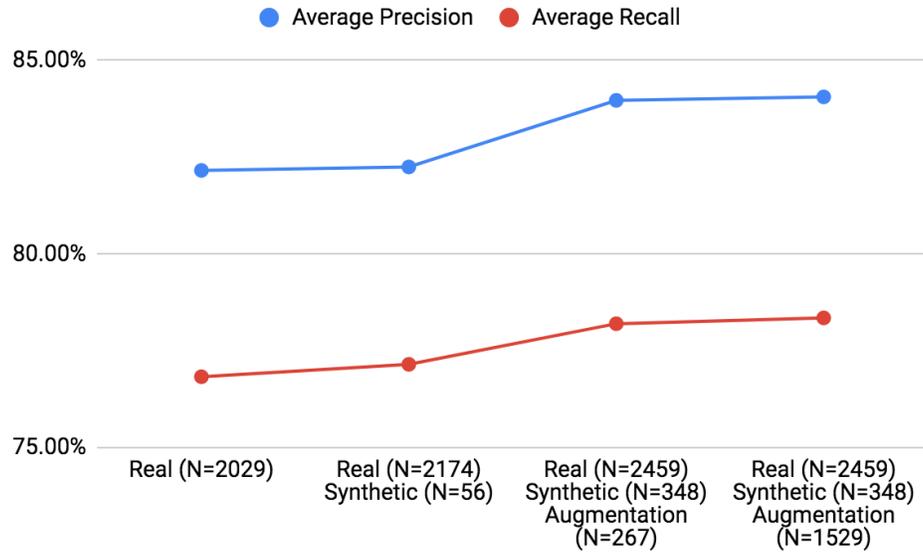

Figure 4. Average precision and recall of different training dataset combinations, including real, synthetic images by 3D game engine, and data augmentation (flip, rotation, zoom in/out).

## ACKNOWLEDGMENT

The research was carried out at the Jet Propulsion Laboratory, California Institute of Technology, under a contract with the National Aeronautics and Space Administration. The research was funded by the U.S. Department of Homeland Security Science and Technology Directorate Next Generation First Responders Apex Program (DHS S&T NGFR) under NASA prime contract NAS7-03001, Task Plan Number 82-106095.